\documentclass[sigconf]{acmart}

\AtBeginDocument{%
  \providecommand\BibTeX{{%
    \normalfont B\kern-0.5em{\scshape i\kern-0.25em b}\kern-0.8em\TeX}}}

\setcopyright{rightsretained}
\copyrightyear{2021}
\acmYear{2021}
\acmDOI{}

\acmConference[MIS2 workshop at KDD 2021]{The Second International MIS2 Workshop: 
Misinformation and Misbehavior Mining on the Web} {2021}{Virtual}
\acmBooktitle{KDD 2021: The Second International MIS2 Workshop: Misinformation and Misbehavior Mining on the Web}
\acmPrice{}
\acmISBN{}

\begin{document}

\title{Detection of Criminal Texts for the Polish State Border Guard}

\author{Artur Nowakowski}
\affiliation{%
  \institution{Adam Mickiewicz University}
  \city{Poznań}
  \country{Poland}}
\email{artur.nowakowski@amu.edu.pl}

\author{Krzysztof Jassem}
\affiliation{%
  \institution{Adam Mickiewicz University}
  \city{Poznań}
  \country{Poland}}
\email{jassem@amu.edu.pl}




\begin{abstract} 
This paper describes research on the detection of Polish criminal texts appearing on the Internet. We carried out experiments to find the best available setup for the efficient classification of unbalanced and noisy data. The best performance was achieved when our model was fine-tuned on a pre-trained Polish-based transformer language model. For the detection task, a large corpus of annotated Internet snippets was collected as training data. We share this dataset and create a new task for the detection of criminal texts using the \textit{Gonito} platform as the benchmark.
\end{abstract}
\maketitle

\section{Introduction}

This paper describes experiments whose goal was to find the best available method for the detection of criminal texts on the Internet. We first verified up-to-date ML methods without pre-training. Then, we experimented with various pre-trained language models, discovering that pre-training significantly improves the performance of a detector.

Our detector is applied in a project entitled “Advanced Internet analysis supporting the detection of criminal groups”\footnote{The project is financed by the Polish National Center for Research and Development.} (the project’s short name is AI Searcher). This 3-year program has the aim of developing a system to support the protection of the eastern border of the European Union by searching the Internet for criminal texts that may be of interest to employees of the Polish State Border Guard. The user scenario is the following: An employee of the State Border Guard types an inquiry into an edit window. The Query Expansion Module expands the inquiry to a set of queries that are semantically related to the inquiry. The Translation Module translates the set of queries into Russian, Ukrainian, and Belarusian. The Crawler searches the Internet to find texts in Polish, Russian, Ukrainian, and Belarusian related to the queries. The Translation Module translates the foreign texts back to Polish. Finally, the Classifier analyzes the texts to return “semaphore information'': a red light for the most likely criminal texts, yellow for suspicious texts, and green for non-criminal texts. Potentially criminal texts are displayed at the beginning of the list of found texts, thus helping the operator in ,,finding a needle in a haystack''. The system user may manually label any text returned by the system as criminal or non-criminal, confirming or contradicting the Classifier's decision. The feedback from the users is stored in the system and may serve as human evaluation of the Classifier. Moreover, these additional annotations will enrich the training data and will be applied to fine-tune the Classifier in the future.

Criminal content of interest to the State Border Guard may relate to the following topics: general smuggling, drugs, medicines, alcohol and cigarettes trafficking, people trafficking, human organs trafficking, weapons and explosives, sex crime, document fraud, or trafficking of stolen cars and machines. There are two difficulties associated with the task: Firstly, the detection of such texts on the Internet with lexical-based or unsupervised methods is not likely to succeed, because the vocabulary of criminal texts does not differ significantly from that of non-criminal content. Secondly, criminal texts are sparse within the legal Internet, so the two classes are strongly unbalanced. 
The contribution of the paper is as follows:

\begin{itemize}
    \item We present a method for the collection of labeled training data for strongly unbalanced classes.
    \item We compare the application of various state-of-the-art transformer language models in a binary classification task based on a highly unbalanced and noisy dataset.
    \item We set up a challenge for the detection of criminal texts by publishing the training and testing data on an evaluation platform.
\end{itemize}

\section{Neural language models}
\label{models}
Neural language models assign probabilities to word sequences using neural networks to calculate the probability values. Recent research shows that one of the most effective architectures of neural networks for building a language model is the transformer \cite{Vaswani2017}. The  architecture, designed initially for a machine translation task, applies the concept of “attention'', which consists in focusing on important parts of the input data.  In the standard unidirectional transformer architecture all tokens attend only to previous (left-to-right models) or subsequent tokens (right-to-left models). In \cite{devlin-etal-2019-bert} the bidirectional architecture BERT (Bidirectional Encoder Representation for Transformers), which uses a “masked language model'', is introduced. The concept consists in masking one of the tokens of the input and guessing its vocabulary identifier based on both left and right contexts.

Initially, neural models were meant to solve tasks at sentence level (such as paraphrasing or natural language inference) or at token level (such as named entity recognition). In \cite{Sun2019} it is shown that the use of the pre-trained BERT model may significantly improve the results of a document-level classification task. 

In this paper we examine the impact of various BERT-like models on the detection of criminal texts on the Internet. The first improvement on the BERT model, named RoBERTa, was made by Facebook researchers \cite{Liu2019}. They show that a few adjustments, such as training the model longer with more data on longer sentences, changing the masking pattern from static to dynamic (where each word sequence is masked not once but in 10 different ways), or dropping an auxiliary factor in the BERT training, namely Next Sentence Prediction (NSP), may significantly improve BERT-based methods in several NLP tasks (such as natural language understanding, question answering or reading comprehension).

Researchers from Hugging Face take a different approach. In \cite{sanh2020distilbert} they show that it is possible to build a smaller language model that achieves similar results on NLP tasks than its larger predecessors. The gain is measured in terms of model size (40\%) and inference time (60\%).

A “large-scale data response'' from Facebook enters the domain of  multilinguality. In \cite{conneau-etal-2020-unsupervised} it is shown that new opportunities open up when the model is trained on multilingual data. The authors' proposition, XML-BERT, outperforms previous solutions on multilingual tasks (such as multilingual question answering or cross-lingual named entity recognition), and at the same time competes with other solutions on monolingual tasks. 

PolBERTa\footnote{https://metatext.io/models/marrrcin-PolBERTa-base-polish-cased-v1} is a RoBERTa-like model pre-trained for the Polish language, which may be easily downloaded and run in a Python environment using the Hugging Face transformers library \cite{wolf-etal-2020-transformers}.

PolBERT\footnote{https://github.com/kldarek/polbert} is a Polish version of the BERT language model, available in two variants (cased and uncased), which can be downloaded and used via the Hugging Face transformers library. The larger (cased) version includes texts from the Polish subset of Open Subtitles, the Polish subset of Paracrawl, the Polish Parliamentary Corpus, and Polish Wikipedia as at February 2020, forming a set of ca. 68 million lines in total.

In \cite{Dadas2020} a 3-month dump from the Polish part of Common Crawls was collected and fused with publicly available Polish text data: the Polish version of Wikipedia, the Polish Parliamentary Corpus, data from smaller projects, individual books, and articles. The authors show that their model outperforms previous solutions in 11 out of 13 NLP tasks defined for the Polish language.

In \cite{mroczkowski-etal-2021-herbert} yet another language model for the Polish language is introduced, named HerBERT. Compared with previous proposals, this adds new publicly available corpora (such as an open portal of Polish Free Readings and texts from an e-commerce portal). The main contribution of this solution is the use of the BERT architecture with dynamic masking, following the training setup of the RoBERTa model and discarding the NSP objective. 

\section{Experiments}
\label{experiments}
\subsection{Data preparation}
We decided to classify Internet pages based on the snippets returned by Internet search engines. The following goals were set for the task of data collection:
\begin{itemize}
\item The corpus of snippets should be large because of the sparsity of the detected class -- we decided on a minimum size of 100,000 items.
\item The focus should be placed on finding texts containing different names of drugs, cigarettes and alcoholic drinks, in order to properly classify ambiguous criminal texts. 
\end{itemize}

To achieve the above goals, we applied the following procedure:
\begin{enumerate}
    \item Manual querying
    
    State Border Guard employees were asked to input their queries of interest into a manual-querying system. The system used the Bing search engine to present a list of snippets relevant to the query. The user labeled the snippets as either interesting (potential criminal content) or non-interesting. 
    
    As a result, an initial corpus of 3,886 snippets was collected, a set of queries of interest to State Border Guard employees was achieved, and a set of rules for manual data labeling was defined based on the State Border Guard employees' choices.   
    \item Automatic collection
    
    We implemented a system, named Data Collector, to automatically collect snippets returned by various search engines in response to queries similar to those asked by the State Border Guard employees. We applied the set of queries assembled in the manual step as input to our Query Expansion Module. The output -- a wider set of potential queries -- was used as the input to the Data Collector module. The tool collected snippets and page content returned by selected search engines (by default: Google, Bing and DuckDuckGo). The number of results pages returned for each query was set to 10. 
    
    \item Annotation
    
    The annotation process was facilitated by a web service dedicated to the task, named AISearcher Tagger. The application enables concurrent work by a group of annotators, whose task is to label each snippet as interesting (potentially criminal) or non-interesting. The decision may be taken based on either the snippet or the whole page content, because of the intuition that snippet content correlates highly with intent to visit the page. Each snippet was labeled by two annotators. The snippet was finally labeled as “interesting'' if it was denoted as such by at least one of the annotators. 
    
The final dataset consisted of 114,432 labeled snippets, among which 2.23\% of snippets were labeled as interesting.
The thematic distribution of classes was unbalanced, because of the State Border Guard's request to place the focus on many different names of drugs which are common in criminal slang. We find that many texts related to the sale and smuggling of drugs are structured similarly to texts related to the sale and smuggling of alcohol, cigarettes, and documents, which also helps to correctly classify these different categories. The thematic distribution of all categories is presented in Table~\ref{table-thematicdist}.

\begin{table}[H]
  \centering
  \caption{Thematic distribution of the dataset}
  \label{table-thematicdist}
\begin{tabular}{ll}
\hline
Theme                  & No. of samples     \\ \hline
Drugs                  & 80,107 (70.01\%) \\
Sale of organs         & 10,301 (9.01\%)  \\
Cigarettes             & 7,244 (6.33\%)    \\
Documents              & 5,175 (4.51\%)    \\
Weapons and explosives & 5,022 (4.39\%)    \\
Alcohol                & 2,904 (2.54\%)    \\
Sex crime              & 2,509 (2.19\%)     \\
Human trafficking      & 1,170 (1.02\%)    \\ \hline
\end{tabular}
\end{table}

The dataset was split into a training set, a validation set and a test set, preserving the distribution of the thematic classes. A total of 92,028 snippets were used for training, 10,570 for validation, and 11,834 for testing. 
\end{enumerate}

\subsection{Training}

We first experimented on the data without the use of any pre-trained language model. We tried two numerical data representations (TF-IDF and Word2Vec) and a series of ML methods (such as Support Vectors Models, Random Forests, XGBoost \cite{Chen:2016:XST:2939672.2939785}, RNN and H2O AutoML \cite{H2OAutoML20}). The best result was achieved by H2O AutoML with the Word2Vec representation. We set this result as a baseline for further experiments with the use of the pre-trained language model. In the next step we experimented with the language models mentioned in section \ref{models}. All of the experiments based on the transformer language models were performed with the aid of the \textit{simpletransformers}\footnote{https://github.com/ThilinaRajapakse/simpletransformers} library.

In the experiments on pre-trained language models, we carried out fine-tuning for a maximum of 5 epochs. We validated every 200 steps with the early stopping set to 10 consecutive validations without improvement. We used Adam \cite{DBLP:journals/corr/KingmaB14} optimizer with the following parameters: \(\beta_1 = 0.99 \), \(\beta_2 = 0.999\), \(\epsilon = 1\mathrm{e}{-8} \). We set the linear decay learning rate scheduling to have a peak value of \(2\mathrm{e}{-5}\). Due to the initial transfer of weights from the already trained models, the warm-up stage was set to 500 steps. We used a batch size of 64. Because of the unbalanced character of the  dataset, we set the class weight of the positive class (snippets labeled as ''interesting'') to 1 and the class weight of the negative class -- to 0.5.

\subsection{Results}
Table \ref{table-results} shows the results of our experiments. The specified evaluation metrics are: F1.0 score, Accuracy (Acc), Precision (Prec) and Recall (Rec). The main evaluation metric was F1.0.

\begin{table}[H]
  \centering
  \caption{Classification performance}
  \label{table-results}
\begin{tabular}{lc|cccc}
\textbf{Model} & \textbf{F1.0} & \textbf{Acc} & \textbf{Prec} & \textbf{Rec}  \\ \hline
H2O AutoML (baseline)     & 50.98 & 97.90 & 67.53 & 40.95  \\ \hline
PolBERTa       & 59.39 & 98.30 & 81.66 & 46.66              \\
XLM-RoBERTa    & 61.77 & 98.32 & 78.81 & 50.79             \\
DistilBERT     & 64.44 & 98.37 & 77.33 & 55.23             \\ 
Polish RoBERTa & 66.17 & 98.45 & 79.20 & 56.82             \\
Polbert        & 67.26 & 98.47 & 78.15 & 59.04            \\
HerBERT        & \textbf{71.07} & 98.52 & 74.13 & 68.25 
\\ \hline
\end{tabular}
\end{table}

\subsection{Conclusion}
    
The experiments show significant improvement of the detector with any of the BERT-based models. The language-specific models handle the task better than the multilingual models. The RoBERTa training setup used in the HerBERT model shows an advantage over the basic BERT setup.

\section{NLP Benchmarks}
We would like to share the data and our experience gathered during the experiments with other researchers. To this end, we have defined an open benchmark\footnote{An NLP benchmark is a platform for evaluating and comparing models for various language processing tasks.} for a task of criminal texts detection, based on our annotated data. In this section, we describe other NLP benchmarks defined for the English and Polish languages. Having presented the background, we describe our benchmark in section \ref{detection}.

\subsection{English benchmarks}
GLUE \cite{wang-etal-2018-glue} is a benchmark consisting of nine NLU (Natural Language Understanding) tasks:
\begin{itemize}
    \item MNLI Multi-Genre Natural Language Inference -- to determine if two sentences are in one of the three relations: entailment, contradiction, neutrality;
    \item RTE Recognizing Textual Entailment -- to determine if two sentences are in one of the two relations: 1) entailment or contradiction, 2) neutrality;
    \item QQP Quora Question Pairs -- to determine if two questions are semantically equivalent;
    \item QNLI Question Natural Language Inference -- to find out if the sentence contains the answer to the question it is paired to;
    \item SST-2 Stanford Sentiment Treebank -- to analyze the sentiment of movie reviews;
    \item CoLA Corpus of Linguistic Acceptability -- to predict if an English sentence is linguistically correct;
    \item STS-B Semantic Textual Similarity Benchmark -- to predict the semantic similarity of two sentences;
    \item MRPC Microsoft Research Paraphrase Corpus -- to determine whether one sentence is a paraphrase of another;
    \item WNLI Winograd Natural Language Inference -- to select, from a list, the reference for a given pronoun in a text. 
    \end{itemize}

SQuAD (Stanford Question Answering Dataset) \cite{rajpurkar-etal-2016-squad} is a collection of over 100,000 pairs of questions and text passages. The task consists in determining whether a given passage contains the answer to a given question.

SWAG (Situations With Adversarial Generations) is the task of completing a sentence with one of four possible continuations. The dataset consists of 113,000 sentence pairs.

\subsection{Polish benchmarks}
The Polish equivalent of GLUE is KLEJ \cite{rybak-etal-2020-klej}. The benchmark consists of nine tasks:
\begin{itemize}
   \item NKLP-NER -- to predict the presence and the type of a named entity in a given sentence;
   \item CDSC-R (Compositional Distributional Semantics Corpus -- Relatedness) -- to determine  the degree of relatedness (from 1 to 5) between two sentences;
   \item CDSC-E (Compositional Distributional Semantics Corpus -- Entailment) -- to determine if two sentences are in one of the three relations: entailment, contradiction, neutrality (as in MNLI);
   \item CBD (Cyberbullying Detection) -- to determine if a Twitter message is a case of cyberbullying;
   \item PolEmo2.0 In Domain -- to predict a sentiment label of a consumer review with the test set belonging to the same domain as the training data;
   \item PolEmo2.0 Out-of-Domain -- to predict a sentiment label of a consumer review with the test set outside the training domain;
   \item DYK (Did you know?) -- to determine if a given Wikipedia article is the answer to a given question;
   \item PSC (Polish Summary Corpus) -- to determine if an automatically generated summary of a given text resembles one of five summaries prepared beforehand by humans;
   \item AR (Allegro Reviews) -- to predict a rating from a given review.
\end{itemize}

A few tasks for processing in the Polish language are presented in \cite{dadas2020evaluation}. They concern semantic analysis, sentence entailment, and topic classification of sentences. The last task (named 8tags) consists in labeling the sentence with one of eight classes: \textit{film, history, food, medicine, motorization, work, sport} and \textit{technology}. 

\section{The Task for Criminal Text Detection}
\label{detection}

Criminal Text Detection is yet another task that we wish to define for the Polish language. The task is published on \textit{Gonito}\footnote{https://gonito.net/} -- an open git-based platform for machine learning competitions. The platform offers several dozen competitions, mostly (but not solely) involving Polish language processing. All of the competitions and their results can be seen without logging in, but in order to participate, an account must be created. Among the most popular tasks published on the \textit{Gonito} platform are:

\begin{itemize}
    \item Challenging America year prediction -- to guess the publication date of an excerpt from ChroniclingAmerica;\footnote{ChroniclingAmerica, the result of a 15-year effort, is a website (https://chroniclingamerica.loc.gov) which provides access to select digitized newspapers.}
    \item Searching for Legal Clauses by Analogy -- to find substrings of a given document which are semantically close to a given sample of another document;
    \item "He Said She Said" -- to determine, for a given text, the gender of its author.  
\end{itemize}

To participate in any of the \textit{Gonito} challenges, it suffices to upload a solution to the platform. The platform automatically evaluates the solution with the metrics appropriate for the task and places the contribution on the leader-board. It is worth noting that all \textit{Gonito} code is open-source,\footnote{https://gitlab.com/filipg/gonito} so that the platform may be reproduced on any server and used for different ML challenges.

On the \textit{Gonito} platform we share the dataset of 114,432 labeled snippets, among which 2.23\% are labeled as interesting.\footnote{The dataset is also available on GitHub: https://github.com/arturnn/criminal-classification-challenge} We split the set into a training set (92,028 snippets), a validation set (10,570 snippets) and a testing set (11,834 snippets). Furthermore, the \textit{Gonito} benchmark provides an automatic evaluation tool, named \textit{GEval} \cite{gralinski-etal-2019-geval}, which returns the F1.0 score for any submitted solution.\footnote{The range of F1.0 scores returned by the GEval tool is from 0 to 1.}

\section{Conclusions and future work}
This paper describes experiments on the detection of Polish criminal texts on the Internet. We present a methodology for the collection of a large corpus of annotated Internet snippets. We analyze the impact of language-specific neural models on the task of unbalanced classification. Our experiments show that using such models for pre-training may significantly improve performance. We discover that the Polish model based on the BERT architecture and the RoBERTa training setup is best suited for highly unbalanced binary classification.

We share our dataset and create a new task for the detection of criminal texts. Furthermore, we share the settings that gave the best results for the task. We use the \textit{Gonito} platform as the benchmark.

We intend to deploy our solution in the AI Searcher system, developed for the needs of the Polish State Border Guard. The organization's employees will have the opportunity to enhance the training dataset. By approving or disapproving the decisions of our detector with a single click, they will contribute to the expansion of the training dataset. The system will then be automatically retrained after a defined number of decisions have been made.

\bibliographystyle{ACM-Reference-Format}
\bibliography{MIS2.bib}


\end{document}